\documentclass[11pt,a4paper]{article}
\usepackage[hyperref]{emnlp-ijcnlp-2019}
\usepackage{times}
\usepackage{latexsym}
\usepackage{url}
\usepackage{times}
\usepackage{latexsym}
\usepackage{amsmath}
\usepackage{amsfonts}
\usepackage{microtype}
\usepackage{xfrac}
\usepackage{mathtools}
\usepackage{tikz}
\usepackage{caption}
\usetikzlibrary{shapes.geometric}
\usetikzlibrary{automata, positioning, arrows}
\usepackage{booktabs}
\usepackage{adjustbox}
\usepackage{xspace}
\usepackage[normalem]{ulem}
\usepackage{subfig}
\usepackage{pifont}
\usepackage{bbm}

\usepackage{xcolor}
\usepackage{algorithm}
\usepackage[noend]{algpseudocode}

\usepackage{cleveref}
\crefname{section}{\S}{\S\S}
\crefname{table}{Tab.}{}
\crefname{figure}{Fig.}{}
\crefname{algorithm}{Alg.}{}
\crefname{equation}{eq.}{}
\crefname{appendix}{App.}{}
\crefformat{section}{\S#2#1#3}  %

\usepackage{todonotes}

\newcommand{\LN}{\text{LN}\xspace}
\newcommand{\Skip}{\text{Skip}\xspace}
\newcommand{\dropout}{\text{Dropout}\xspace}
\newcommand{\FF}{\text{FF}\xspace}
\newcommand{\SA}{\text{MHSA}\xspace}

\newcommand{\gelu}{\text{GELU}\xspace}
\newcommand{\concat}{\text{Concat}\xspace}
\newcommand{\yy}{\mathbf{y}}
\newcommand{\xx}{\mathbf{x}}
\newcommand{\bb}{\mathbf{b}}
\newcommand{\ptype}{p_\text{type}}
\newcommand{\ptoken}{p_\text{token}}
\newcommand{\WW}{\mathbf{W}}
\newcommand{\calR}{\mathbb{R}}
\newcommand{\concurrent}{\textdagger\xspace}
\newcommand{\bitext}{\ensuremath{\spadesuit}\xspace}
\newcommand{\devselect}{\ensuremath{\diamondsuit}\xspace}

\newcommand\blfootnote[1]{%
  \begingroup
  \renewcommand\thefootnote{}\footnote{#1}%
  \addtocounter{footnote}{-1}%
  \endgroup
}

\aclfinalcopy %

\newcommand{\insertLangTable}{
\begin{table*}[t]
\centering
\begin{adjustbox}{width=2\columnwidth}
\begin{tabular}{ccccc ccccc ccccc ccccc ccccc ccccc ccccc ccccc}
 \toprule
 & ar & bg & ca & cs & da & de & el & en & es & et & fa & fi & fr & he & hi & hr & hu & id & it & ja & ko & la & lv & nl & no & pl & pt & ro & ru & sk & sl & sv & sw & th & tr & uk & ur & vi & zh \\  \midrule
MLDoc &  &  &  &  &  & $\checkmark$ &  & $\checkmark$ & $\checkmark$ &  &  &  & $\checkmark$ &  &  &  &  &  & $\checkmark$ & $\checkmark$ &  &  &  &  &  &  &  &  & $\checkmark$ &  &  &  &  &  &  &  &  &  & $\checkmark$ \\
NLI & $\checkmark$ & $\checkmark$ &  &  &  & $\checkmark$ & $\checkmark$ & $\checkmark$ & $\checkmark$ &  &  &  & $\checkmark$ &  & $\checkmark$ &  &  &  &  &  &  &  &  &  &  &  &  &  & $\checkmark$ &  &  &  & $\checkmark$ & $\checkmark$ & $\checkmark$ &  & $\checkmark$ & $\checkmark$ & $\checkmark$ \\
NER &  &  &  &  &  & $\checkmark$ &  & $\checkmark$ & $\checkmark$ &  &  &  &  &  &  &  &  &  &  &  &  &  &  & $\checkmark$ &  &  &  &  &  &  &  &  &  &  &  &  &  &  & $\checkmark$ \\
POS &  & $\checkmark$ &  &  & $\checkmark$ & $\checkmark$ &  & $\checkmark$ & $\checkmark$ &  & $\checkmark$ &  &  &  &  &  & $\checkmark$ &  & $\checkmark$ &  &  &  &  & $\checkmark$ &  & $\checkmark$ & $\checkmark$ & $\checkmark$ &  & $\checkmark$ & $\checkmark$ & $\checkmark$ &  &  &  &  &  &  &  \\
Parsing & $\checkmark$ & $\checkmark$ & $\checkmark$ & $\checkmark$ & $\checkmark$ & $\checkmark$ &  & $\checkmark$ & $\checkmark$ & $\checkmark$ &  & $\checkmark$ & $\checkmark$ & $\checkmark$ & $\checkmark$ & $\checkmark$ &  & $\checkmark$ & $\checkmark$ & $\checkmark$ & $\checkmark$ & $\checkmark$ & $\checkmark$ & $\checkmark$ & $\checkmark$ & $\checkmark$ & $\checkmark$ & $\checkmark$ & $\checkmark$ & $\checkmark$ & $\checkmark$ & $\checkmark$ &  &  &  & $\checkmark$ &  &  & $\checkmark$ \\ \bottomrule

\end{tabular}
\end{adjustbox}
\caption{The 39 languages used in the 5 tasks.}
\label{tab:lang}
\end{table*}
}

\newcommand{\insertMLDocTable}{
\begin{table*}[t]
\begin{center}
\resizebox{.7\linewidth}{!}{
\begin{tabular}[b]{l|ccccc ccc|c}
\toprule
& en & de & zh & es & fr & it & ja & ru & Average \\
\midrule
\multicolumn{10}{l}{\it In language supervised learning} \\
\midrule
\citet{schwenk-li-2018-corpus} & 92.2 & 93.7 & 87.3 & 94.5 & 92.1 & 85.6 & 85.4 & 85.7 & 89.5 \\
mBERT & 94.2 & 93.3 & 89.3 & 95.7 & 93.4 & 88.0 & 88.4 & 87.5 & 91.2 \\
\midrule
\multicolumn{10}{l}{\it Zero-shot cross-lingual transfer} \\
\midrule
\citet{schwenk-li-2018-corpus} & \underline{92.2} & \underline{81.2} & \underline{74.7} & 72.5 & 72.4 & \textbf{69.4} & \textbf{67.6} & 60.8 & 73.9 \\
\citet{artetxe2018massively} \bitext \concurrent & 89.9 & \textbf{84.8} & 71.9 & \textbf{77.3} & \textbf{78.0} & \textbf{69.4} & \underline{60.3} & \underline{67.8} & \textbf{74.9} \\
mBERT & \textbf{94.2} & 80.2 & \textbf{76.9} & \underline{72.6} & \underline{72.6} & \underline{68.9} & 56.5 & \textbf{73.7} & \underline{74.5} \\
\bottomrule
\end{tabular}
}
\caption{MLDoc experiments. \bitext denotes the model is pretrained with bitext, and \concurrent denotes concurrent work. Bold and underline denote best and second best.}
\label{tab:mldoc}
\end{center}
\end{table*}
}

\newcommand{\insertXNLITable}{
\begin{table*}[]
\begin{center}
\resizebox{1\linewidth}{!}{
\begin{tabular}[b]{l|ccccccccccccccc|c}
\toprule
& en & fr & es & de & el & bg & ru & tr & ar & vi & th & zh & hi & sw & ur & Average \\
\midrule
\multicolumn{16}{l}{\it Pseudo supervision with machine translated training data from English to target language} \\
\midrule
\citet{lample2019cross} (MLM+TLM) \bitext \concurrent & 85.0 & 80.2 & 80.8 & 80.3 & 78.1 & 79.3 & 78.1 & 74.7 & 76.5 & 76.6 & 75.5 & 78.6 & 72.3 & 70.9 & 63.2 & 76.7 \\
mBERT & 82.1 & 76.9 & 78.5 & 74.8 & 72.1 & 75.4 & 74.3 & 70.6 & 70.8 & 67.8 & 63.2 & 76.2 & 65.3 & 65.3 & 60.6 & 71.6 \\
\midrule
\multicolumn{16}{l}{\it Zero-shot cross-lingual transfer} \\ 
\midrule
\citet{conneau-etal-2018-xnli} (X-LSTM) \bitext \devselect & 73.7 & 67.7 & 68.7 & 67.7 & 68.9 & 67.9 & 65.4 & 64.2 & 64.8 & 66.4 & 64.1 & 65.8 & 64.1 & 55.7 & 58.4 & 65.6 \\
\citet{artetxe2018massively} \bitext \concurrent & 73.9 & 71.9 & 72.9 & 72.6 & 73.1 & 74.2 & 71.5 & 69.7 & 71.4 & 72.0 & 69.2 & 71.4 & 65.5 & 62.2 & 61.0 & 70.2 \\
\citet{lample2019cross} (MLM+TLM) \bitext \devselect \concurrent & 85.0 & 78.7 & 78.9 & 77.8 & 76.6 & 77.4 & 75.3 & 72.5 & 73.1 & 76.1 & 73.2 & 76.5 & 69.6 & 68.4 & 67.3 & 75.1 \\
\midrule
\citet{lample2019cross} (MLM) \devselect \concurrent & 83.2 & 76.5 & 76.3 & 74.2 & 73.1 & 74.0 & 73.1 & 67.8 & 68.5 & 71.2 & 69.2 & 71.9 & 65.7 & 64.6 & 63.4 & 71.5 \\
mBERT & 82.1 & 73.8 & 74.3 & 71.1 & 66.4 & 68.9 & 69.0 & 61.6 & 64.9 & 69.5 & 55.8 & 69.3 & 60.0 & 50.4 & 58.0 & 66.3 \\
\bottomrule
\end{tabular}
}
\caption{XNLI experiments. \bitext denotes the model is pretrained with cross-lingual signal including bitext or bilingual dictionary, \concurrent denotes concurrent work, and \devselect denotes model selection with target language dev set.}
\label{tab:xnli}
\end{center}
\end{table*}
}

\newcommand{\insertNerTable}{
\begin{table}[]
\begin{center}
\resizebox{\linewidth}{!}{
\begin{tabular}[b]{l|ccccc|c}
\toprule
& en & nl & es & de & zh & Average (-en,-zh) \\
\midrule
\multicolumn{7}{l}{\it In language supervised learning} \\
\midrule
\citet{xie-etal-2018-neural} & - & 86.40 & 86.26 & 78.16 & - & 83.61 \\
mBERT & 91.97 & 90.94 & 87.38 & 82.82 & 93.17 & 87.05 \\
\midrule
\multicolumn{7}{l}{\it Zero-shot cross-lingual transfer} \\
\midrule
\citet{xie-etal-2018-neural} \devselect & - & 71.25 & 72.37 & 57.76 & - & 67.13 \\
mBERT & 91.97 & \textbf{77.57} & \textbf{74.96} & \textbf{69.56} & 51.90 & \textbf{74.03} \\
\bottomrule
\end{tabular}
}
\caption{NER tagging experiments. \devselect denotes model selection with target language dev set.}
\label{tab:ner}
\end{center}
\end{table}
}

\newcommand{\insertPosTable}{
\begin{table*}[]
\begin{center}
\resizebox{\linewidth}{!}{
\begin{tabular}[b]{l|ccccc ccccc ccccc|c}
\toprule
lang & bg & da & de & en & es & fa & hu & it & nl & pl & pt & ro & sk & sl & sv & Average (-en) \\
\midrule
\multicolumn{10}{l}{\it In language supervised learning} \\
\midrule
mBERT & 99.0 & 97.9 & 95.2 & 97.1 & 97.1 & 97.8 & 96.9 & 98.7 & 92.1 & 98.5 & 98.3 & 97.8 & 97.0 & 98.9 & 98.4 & 97.4 \\
\midrule
\multicolumn{10}{l}{\it Low resource cross-lingual transfer} \\
\midrule
\citet{kim-etal-2017-cross} (1280) & 95.7 & 94.3 & 90.7 & - & 93.4 & 94.8 & 94.5 & 95.9 & 85.8 & 92.1 & 95.5 & 94.2 & 90.0 & 94.1 & 94.6 & 93.3 \\
\citet{kim-etal-2017-cross} (320) & 92.4 & 90.8 & 89.7 & - & 90.9 & 91.8 & 90.7 & 94.0 & 82.2 & 85.5 & 94.2 & 91.4 & 83.2 & 90.6 & 90.7 & 89.9 \\
\midrule
\multicolumn{10}{l}{\it Zero-shot cross-lingual transfer} \\
\midrule
mBERT & 87.4 & 88.3 & 89.8 & 97.1 & 85.2 & 72.8 & 83.2 & 84.7 & 75.9 & 86.9 & 82.1 & 84.7 & 83.6 & 84.2 & 91.3 & 84.3 \\

\bottomrule
\end{tabular}
}
\caption{POS tagging. \citet{kim-etal-2017-cross} use small amounts of training data in the target language.}
\label{tab:pos}
\end{center}
\end{table*}
}

\newcommand{\insertParsingTable}{
\begin{table}[t]
\begin{center}
\resizebox{\linewidth}{!}{
\begin{tabular}[b]{l ccccc}
\toprule
 & Dist & mBERT(S) & Baseline(Z) & mBERT(Z) & mBERT(Z+POS) \\
\midrule
en & 0.00 & 91.5/81.3 & 90.4/\textbf{88.4} & \underline{91.5}/81.3 & \textbf{91.8}/\underline{82.2} \\
no & 0.06 & 93.6/85.9 & \underline{80.8}/\textbf{72.8} & 80.6/68.9 & \textbf{82.7}/\underline{72.1} \\
sv & 0.07 & 91.2/83.1 & 81.0/\underline{73.2} & \underline{82.5}/71.2 & \textbf{84.3}/\textbf{73.7} \\
fr & 0.09 & 91.7/85.4 & 77.9/\underline{72.8} & \underline{82.7}/72.7 & \textbf{83.8}/\textbf{76.2} \\
pt & 0.09 & 93.2/87.2 & 76.6/\textbf{67.8} & \underline{77.1}/64.0 & \textbf{78.3}/\underline{66.9} \\
da & 0.10 & 89.5/81.9 & 76.6/\underline{67.9} & \underline{77.4}/64.7 & \textbf{79.3}/\textbf{68.1} \\
es & 0.12 & 92.3/86.5 & 74.5/\underline{66.4} & \underline{78.1}/64.9 & \textbf{79.0}/\textbf{68.9} \\
it & 0.12 & 94.8/88.7 & 80.8/\underline{75.8} & \underline{84.6}/74.4 & \textbf{86.0}/\textbf{77.8} \\
ca & 0.13 & 94.3/89.5 & 73.8/\underline{65.1} & \underline{78.1}/64.6 & \textbf{79.0}/\textbf{67.9} \\
hr & 0.13 & 92.4/83.8 & 61.9/52.9 & \textbf{80.7}/\underline{65.8} & \underline{80.4}/\textbf{68.2} \\
pl & 0.13 & 94.7/79.9 & 74.6/\underline{62.2} & \underline{82.8}/59.4 & \textbf{85.7}/\textbf{65.4} \\
sl & 0.13 & 88.0/77.8 & 68.2/\underline{56.5} & \underline{72.6}/51.4 & \textbf{75.9}/\textbf{59.2} \\
uk & 0.13 & 90.6/83.4 & 60.1/52.3 & \textbf{76.7}/\underline{60.0} & \underline{76.5}/\textbf{65.5} \\
bg & 0.14 & 95.2/85.5 & 79.4/\textbf{68.2} & \underline{83.3}/62.3 & \textbf{84.4}/\underline{68.1} \\
cs & 0.14 & 94.2/86.6 & 63.1/53.8 & \underline{76.6}/\underline{58.7} & \textbf{77.4}/\textbf{63.6} \\
de & 0.14 & 86.1/76.5 & 71.3/61.6 & \underline{80.4}/\underline{66.3} & \textbf{83.5}/\textbf{71.2} \\
he & 0.14 & 91.9/83.6 & 55.3/48.0 & \textbf{67.5}/\underline{48.4} & \underline{67.0}/\textbf{54.3} \\
nl & 0.14 & 94.0/85.0 & 68.6/60.3 & \underline{78.0}/\underline{64.8} & \textbf{79.9}/\textbf{67.1} \\
ru & 0.14 & 94.7/88.0 & 60.6/51.6 & \textbf{73.6}/\underline{58.5} & \underline{73.2}/\textbf{61.5} \\
ro & 0.15 & 92.2/83.2 & 65.1/54.1 & \textbf{77.0}/\underline{58.5} & \underline{76.9}/\textbf{62.6} \\
id & 0.17 & 86.3/75.4 & 49.2/43.5 & \textbf{62.6}/\underline{45.6} & \underline{59.8}/\textbf{48.6} \\
sk & 0.17 & 93.8/83.3 & 66.7/58.2 & \underline{82.7}/\underline{63.9} & \textbf{82.9}/\textbf{67.8} \\
lv & 0.18 & 87.3/75.3 & \textbf{70.8}/\textbf{49.3} & 66.0/41.4 & \underline{70.4}/\underline{48.5} \\
et & 0.20 & 88.8/79.7 & 65.7/\underline{44.9} & \underline{66.9}/44.3 & \textbf{70.8}/\textbf{50.7} \\
fi & 0.20 & 91.3/81.8 & 66.3/\underline{48.7} & \underline{68.4}/47.5 & \textbf{71.4}/\textbf{52.5} \\
zh* & 0.23 & 88.3/81.2 & 42.5/25.1 & \textbf{53.8}/\underline{26.8} & \underline{53.4}/\textbf{29.0} \\
ar & 0.26 & 87.6/80.6 & 38.1/28.0 & \underline{43.9}/\underline{28.3} & \textbf{44.7}/\textbf{32.9} \\
la & 0.28 & 85.2/73.1 & \underline{48.0}/\textbf{35.2} & 47.9/26.1 & \textbf{50.9}/\underline{32.2} \\
ko & 0.33 & 86.0/74.8 & 34.5/16.4 & \textbf{52.7}/\underline{27.5} & \underline{52.3}/\textbf{29.4} \\
hi & 0.40 & 94.8/86.7 & 35.5/26.5 & \underline{49.8}/\underline{33.2} & \textbf{58.9}/\textbf{44.0} \\
ja* & 0.49 & 94.2/87.4 & 28.2/\underline{20.9} & \underline{36.6}/15.7 & \textbf{41.3}/\textbf{30.9} \\
\midrule
AVER & 0.17 & 91.3/82.6 & 64.1/53.8 & \underline{71.4}/\underline{54.2} & \textbf{73.0}/\textbf{58.9} \\
\bottomrule
\end{tabular}
}
\caption{Dependency parsing results by language (UAS/LAS). * denotes delexicalized parsing in the baseline. S and Z denotes supervised learning and zero-shot transfer. Bold and underline denotes best and second best. We order the languages by word order distance to English.}
\label{tab:parsing}
\end{center}
\end{table}
}

\title{Beto, Bentz, Becas: The Surprising Cross-Lingual Effectiveness of BERT}
  
\author{Shijie Wu \and Mark Dredze \\
Department of Computer Science \\
Johns Hopkins University \\
{\tt shijie.wu@jhu.edu, mdredze@cs.jhu.edu}
}

\date{}

\begin{document}
\maketitle
\begin{abstract}
Pretrained contextual representation models \cite{peters-etal-2018-deep,devlin-etal-2019-bert} have pushed forward the state-of-the-art on many NLP tasks. A new release of BERT \cite{multilingualBERTmd} includes a model simultaneously pretrained on 104 languages with impressive performance for zero-shot cross-lingual transfer on a natural language inference task. This paper explores the broader cross-lingual potential of mBERT (multilingual) as a zero-shot language transfer model on 5 NLP tasks covering a total of 39 languages from various language families: NLI, document classification, NER, POS tagging, and dependency parsing. We compare mBERT with the best-published methods for zero-shot cross-lingual transfer and find mBERT competitive on each task.
Additionally, we investigate the most effective strategy for utilizing mBERT in this manner, determine to what extent mBERT generalizes away from language-specific features, and measure factors that influence cross-lingual transfer.\blfootnote{Code is available at \url{https://github.com/shijie-wu/crosslingual-nlp}}

\end{abstract}

\section{Introduction}

Pretrained language representations with self-supervised objectives have become standard in a variety of NLP tasks \cite{peters-etal-2018-deep,howard-ruder-2018-universal,radford2018improving,devlin-etal-2019-bert}, including sentence-level classification \cite{wang-etal-2018-glue}, sequence tagging (e.g. NER) \cite{tjong-kim-sang-de-meulder-2003-introduction} and SQuAD question answering \cite{rajpurkar-etal-2016-squad}. Self-supervised objectives include language modeling, the cloze task \cite{taylor1953cloze} and next sentence classification. These objectives continue key ideas in word embedding objectives like CBOW and skip-gram \cite{mikolov2013efficient}. 

At the same time, cross-lingual embedding models have reduced the amount of cross-lingual supervision required to produce reasonable models; \newcite{conneau2017word,artetxe-etal-2018-robust} use identical strings between languages as a pseudo bilingual dictionary to learn a mapping between monolingual-trained embeddings.
Can jointly training contextual embedding models over multiple languages without explicit mappings produce an effective cross-lingual representation?
Surprisingly, the answer is (partially) yes. BERT, a recently introduced pretrained model \cite{devlin-etal-2019-bert}, offers a multilingual model (mBERT) pretrained on concatenated Wikipedia data for 104 languages {\em without any cross-lingual alignment} \cite{multilingualBERTmd}. mBERT does surprisingly well compared to cross-lingual word embeddings on zero-shot cross-lingual transfer in XNLI \cite{conneau-etal-2018-xnli}, a natural language inference dataset. {\bf Zero-shot cross-lingual transfer}, also known as single-source transfer, refers \textit{trains and selects} a model in a source language, often a high resource language, then transfers directly to a target language.

While XNLI results are promising, the question remains: does mBERT learn a cross-lingual space that supports zero-shot transfer? We evaluate mBERT as a zero-shot cross-lingual transfer model on five different NLP tasks: natural language inference, document classification, named entity recognition, part-of-speech tagging, and dependency parsing. We show that it achieves competitive or even state-of-the-art performance with the recommended fine-tune all parameters scheme \cite{devlin-etal-2019-bert}. Additionally, we explore different fine-tuning and feature extraction schemes and demonstrate that with parameter freezing, we further outperform the suggested fine-tune all approach. Furthermore, we explore
the extent to which mBERT generalizes away from a specific language by measuring accuracy on language ID using each layer of mBERT. Finally, we show how subword tokenization influences transfer by measuring subword overlap between languages.

\section{Background}

\paragraph{(Zero-shot) Cross-lingual Transfer}

Cross-lingual transfer learning is a type of transductive transfer learning with different source and target domain \citep{pan2010survey}. A cross-lingual representation space is assumed to perform the cross-lingual transfer. Before the widespread use of cross-lingual word embeddings, task-specific models assumed coarse-grain representation like part-of-speech tags, in support of a delexicalized parser \cite{zeman-resnik-2008-cross}. More recently cross-lingual word embeddings have been used in conjunction with task-specific neural architectures for tasks like named entity recognition \cite{xie-etal-2018-neural}, part-of-speech tagging \cite{kim-etal-2017-cross} and dependency parsing \cite{ahmad-etal-2019-difficulties}.

\paragraph{Cross-lingual Word Embeddings.}
The quality of the cross-lingual space is essential for zero-shot cross-lingual transfer. \newcite{ruder2017survey} surveys methods for learning cross-lingual word embeddings by either joint training or post-training mappings of monolingual embeddings.
\newcite{conneau2017word} and \newcite{artetxe-etal-2018-robust} first show two monolingual embeddings can be aligned by learning an orthogonal mapping with only identical strings as an initial heuristic bilingual dictionary.

\paragraph{Contextual Word Embeddings}
ELMo \cite{peters-etal-2018-deep}, a deep LSTM \cite{hochreiter1997long} pretrained with a language modeling objective, learns contextual word embeddings. This contextualized representation outperforms stand-alone word embeddings, e.g. Word2Vec \cite{mikolov2013distributed} and Glove \cite{pennington-etal-2014-glove}, with the same task-specific architecture in various downstream tasks. Instead of taking the representation from a pretrained model, GPT \cite{radford2018improving} and \newcite{howard-ruder-2018-universal} also fine-tune all the parameters of the pretrained model for a specific task. Also, GPT uses a transformer encoder \cite{vaswani2017attention} instead of an LSTM and jointly fine-tunes with the language modeling objective. \newcite{howard-ruder-2018-universal} propose another fine-tuning strategy by using a different learning rate for each layer with learning rate warmup and gradual unfreezing.

Concurrent work by \newcite{lample2019cross} incorporates bitext into BERT by training on pairs of parallel sentences.
\newcite{schuster-etal-2019-cross} aligns pretrained ELMo of different languages by learning an orthogonal mapping and shows strong zero-shot and few-shot cross-lingual transfer performance on dependency parsing with 5 Indo-European languages.
Similar to multilingual BERT, \newcite{mulcaire-etal-2019-polyglot} trains a single ELMo on distantly related languages and shows mixed results as to the benefit of pretaining.

Parallel to our work, \newcite{pires-etal-2019-multilingual} shows mBERT has good zero-shot cross-lingual transfer performance on NER and POS tagging. They show how subword overlap and word ordering effect mBERT transfer performance. Additionally, they show mBERT can find translation pairs and works on code-switched POS tagging.
In comparison, our work looks at a larger set of NLP tasks including dependency parsing and ground the mBERT performance against previous state-of-the-art on zero-shot cross-lingual transfer. We also probe mBERT in different ways and show a more complete picture of the cross-lingual effectiveness of mBERT.

\section{Multilingual BERT}
\label{sec:mBERT}

\paragraph{BERT} \cite{devlin-etal-2019-bert} is a deep contextual representation based on a series of transformers trained by a self-supervised objective. One of the main differences between BERT and related work like ELMo and GPT is that BERT is trained by the Cloze task \cite{taylor1953cloze}, also referred to as masked language modeling, instead of right-to-left or left-to-right language modeling. This allows the model to freely encode information from both directions in each layer. 
Additionally, BERT also optimizes a next sentence classification objective. At training time, 50\% of the paired sentences are consecutive sentences while the rest of the sentences are paired randomly.
Instead of operating on words, BERT uses a subword vocabulary with WordPiece \cite{wu2016google}, a data-driven approach to break up a word into subwords.

\paragraph{Fine-tuning BERT}
BERT shows strong performance by fine-tuning the transformer encoder followed by a softmax classification layer on various sentence classification tasks. A sequence of shared softmax classifications produces sequence tagging models for tasks like NER. Fine-tuning usually takes 3 to 4 epochs with a relatively small learning rate, for example, 3e-5.

\paragraph{Multilingual BERT} mBERT \cite{multilingualBERTmd} follows the same model architecture and training procedure as BERT, except with data from Wikipedia in 104 languages. Training makes no use of explicit cross-lingual signal, e.g. pairs of words, sentences or documents linked across languages.
In mBERT, the WordPiece modeling strategy allows the model to share embeddings across languages. For example, ``DNA'' has a similar meaning even in distantly related languages like English and Chinese \footnote{``DNA'' indeed appears in the vocabulary of mBERT as a stand-alone lexicon.}. To account for varying sizes of Wikipedia training data in different languages, training uses a heuristic to subsample or oversample words when running WordPiece as well as sampling a training batch, random words for cloze and random sentences for next sentence classification. 

\paragraph{Transformer} For completeness, we describe the Transformer used by BERT. Let $\xx$, $\yy$ be a sequence of subwords from a sentence pair. A special token \texttt{[CLS]} is prepended to $\xx$ and \texttt{[SEP]} is appended to both $\xx$ and $\yy$. The embedding is obtained by
\begin{align*}
\hat{h}^0_i &= E(x_i) + E(i) + E(\mathbbm{1}_\xx)  \\
\hat{h}^0_{j+|\mathbf{x}|} &= E(y_j) + E(j+|\mathbf{x}|) + E(\mathbbm{1}_\yy)  \\
h_\cdot^{0} &= \dropout(\LN(\hat{h}_\cdot^{0})) 
\end{align*}
where $E$ is the embedding function and $\LN$ is layer normalization \cite{ba2016layer}. $M$ transformer blocks are followed by the embeddings. In each transformer block,
\begin{align*}
h^{i+1}_{\cdot} &= \Skip(\FF, \Skip(\SA, h^i_{\cdot}))  \\
\Skip(f, h) &= \LN(h + \dropout(f(h)))  \\
\FF(h) &= \gelu(h \WW_1^\top + \bb_1) \WW_2^\top + \bb_2 
\end{align*}
where $\gelu$ is an element-wise activation function \cite{hendrycks2016bridging}. In practice, $h^i \in \calR^{(|\xx| + |\yy|)\times d_h}$, $\WW_1 \in \calR^{4d_h \times d_h}$, $\bb_1 \in \calR^{4d_h}$, $\WW_2 \in \calR^{d_h \times 4d_h}$, and $\bb_2 \in \calR^{d_h}$. $\SA$ is the multi-heads self-attention function. We show how one new position $\hat{h}_i$ is computed.
\begin{align*}
[\cdots,\hat{h}_i,\cdots] &=\SA([h_1, \cdots, h_{|\xx|+|\yy|}]) \\
&= \WW_o \concat(h^1_i, \cdots, h^N_i) + \bb_o
\end{align*}
In each attention, referred to as attention head,
\begin{align*}
h^j_i &= \sum^{|\xx|+|\yy|}_{k=1} \dropout(\alpha^{(i,j)}_k) \WW^j_V h_k \\
\alpha^{(i,j)}_k &= \frac{\exp \frac {(\WW^j_Q h_i)^\top \WW^j_K h_k} {\sqrt{d_h/N}} }
{\sum^{|\xx|+|\yy|}_{k'=1} \exp \frac {(\WW^j_Q h_i)^\top \WW^j_K h_{k'}} {\sqrt{d_h/N}}}
\end{align*}
where $N$ is the number of attention heads, $h^j_i \in \calR^{d_h / N}$, $\WW_o \in \calR^{d_h\times d_h}$, $\bb_o \in \calR^{d_h}$, and $\WW^j_Q,\WW^j_K,\WW^j_V \in \calR^{d_h/N\times d_h}$.

\section{Tasks}

\insertLangTable
Does mBERT learn a cross-lingual representation, or does it produce a representation for each language in its own embedding space?
We consider five tasks in the zero-shot transfer setting. We assume labeled training data for each task in English, and transfer the trained model to a target language. We select a range of different tasks: document classification, natural language inference, named entity recognition, part-of-speech tagging, and dependency parsing. We cover zero-shot transfer from English to 38 languages in the 5 different tasks as shown in \cref{tab:lang}. In this section, we describe the tasks as well as task-specific layers.

\subsection{Document Classification}\label{sec:mldoc}
We use MLDoc \cite{schwenk-li-2018-corpus}, a balanced subset of the Reuters corpus covering 8 languages for document classification. The 4-way topic classification task decides between CCAT (Corporate/Industrial), ECAT (Economics), GCAT (Government/Social), and MCAT (Markets). We only use the first two sentences\footnote{We only use the first sentence if the document only contains one sentence. Documents are segmented into sentences with NLTK \cite{perkins2014python}.} of a document for classification due to memory constraint. The sentence pairs are provided to the mBERT encoder. The task-specific classification layer is a linear function mapping $h^{12}_0 \in \calR^d_h$ into $\calR^4$, and a softmax is used to get class distribution. We evaluate by classification accuracy.

\subsection{Natural Language Inference}
We use XNLI \cite{conneau-etal-2018-xnli} which cover 15 languages for natural language inference. The 3-way classification includes entailment, neutral, and contradiction given a pair of sentences. We feed a pair of sentences directly into mBERT and the task-specific classification layer is the same as \cref{sec:mldoc}. We evaluate by classification accuracy.

\subsection{Named Entity Recognition}\label{sec:ner}

We use the CoNLL 2002 and 2003 NER shared tasks \cite{tjong-kim-sang-2002-introduction,tjong-kim-sang-de-meulder-2003-introduction} (4 languages) and a Chinese NER dataset \cite{levow-2006-third}. The labeling scheme is BIO with 4 types of named entities. We add a linear classification layer with softmax to obtain word-level predictions. Since mBERT operates at the subword-level while the labeling is word-level, if a word is broken into multiple subwords, we mask the prediction of non-first subwords.
NER is evaluated by F1 of predicted entity (F1). Note we use a simple post-processing heuristic to obtain a valid span.

\subsection{Part-of-Speech Tagging}

We use a subset of Universal Dependencies (UD) Treebanks (v1.4) \cite{ud1.4}, which cover 15 languages, following the setup of \newcite{kim-etal-2017-cross}. The task-specific labeling layer is the same as \cref{sec:ner}. POS tagging is evaluated by the accuracy of predicted POS tags (ACC).

\subsection{Dependency parsing}

Following the setup of \newcite{ahmad-etal-2019-difficulties}, we use a subset of Universal Dependencies (UD) Treebanks (v2.2) \cite{ud2.2}, which includes 31 languages. Dependency parsing is evaluated by unlabelled attachment score (UAS) and labeled attachment score (LAS) \footnote{Punctuations (PUNCT) and symbols (SYM) are excluded.}. We only predict the coarse-grain dependency label following Ahmad et al. We use the model of \newcite{dozat2016deep}, a graph-based parser as a task-specific layer. Their LSTM encoder is replaced by mBERT. Similar to \cref{sec:ner}, we only take the representation of the first subword of each word. We use masking to prevent the parser from operating on non-first subwords.

\section{Experiments}

We use the base cased multilingual BERT, which has $N=12$ attention heads and $M=12$ transformer blocks. The dropout probability is 0.1 and $d_h$ is 768. The model has 179M parameters with about 120k vocabulary.

\paragraph{Training} For each task, no preprocessing is performed except tokenization of words into subwords with WordPiece. We use Adam \cite{kingma2014adam} for fine-tuning with $\beta_1$ of 0.9, $\beta_2$ of 0.999 and L2 weight decay of 0.01. We warm up the learning rate over the first 10\% of batches and linearly decay the learning rate.

\paragraph{Maximum Subwords Sequence Length} At training time, we limit the length of subwords sequence to 128 to fit in a single GPU for all tasks. For NER and POS tagging, we additionally use the sliding window approach. After the first window, we keep the last 64 subwords from the previous window as context. In other words, for a non-first window, only (up to) 64 new subwords are added for prediction. At evaluation time, we follow the same approach as training time except for parsing. We threshold the sentence length to 140 words, including words and punctuation, following \newcite{ahmad-etal-2019-difficulties}. In practice, the maximum subwords sequence length is the number of subwords of the first 140 words or 512, whichever is smaller.

\paragraph{Hyperparameter Search and Model Selection} We select the best hyperparameters by searching a combination of batch size, learning rate and the number of fine-tuning epochs with the following range: learning rate $\{2\times 10^{-5},3\times 10^{-5},5\times 10^{-5}\}$; batch size $\{16, 32\}$; number of epochs: $\{3, 4\}$. Note the best hyperparameters and model are selected by development performance in \textit{English}.

\subsection{Question \#1: Is mBERT Multilingual?}\label{sec:exp1}

\insertMLDocTable

\paragraph{MLDoc} We include two strong baselines. \citet{schwenk-li-2018-corpus} use MultiCCA, multilingual word embeddings trained with a bilingual dictionary \cite{ammar2016massively}, and convolution neural networks. Concurrent to our work, \citet{artetxe2018massively} use bitext between English/Spanish and the rest of languages to pretrain a multilingual sentence representation with a sequence-to-sequence model where the decoder only has access to a max-pooling of the encoder hidden states.

mBERT outperforms (\cref{tab:mldoc}) multilingual word embeddings and performs comparably with a multilingual sentence representation, even though mBERT does not have access to bitext. Interestingly, mBERT outperforms \citet{artetxe2018massively} in distantly related languages like Chinese and Russian and under-performs in closely related Indo-European languages.

\insertXNLITable

\paragraph{XNLI} We include three strong baselines, \citet{artetxe2018massively} and \citet{lample2019cross} are concurrent to our work. \citet{lample2019cross} with MLM is similar to mBERT; the main difference is that it only trains with the 15 languages of XNLI, has 249M parameters (around 40\% more than mBERT), and MLM+TLM also uses bitext as training data \footnote{They also use language embeddings as input and exclude the next sentence classification objective}. \citet{conneau-etal-2018-xnli} use supervised multilingual word embeddings with an LSTM encoder and max-pooling. After an English encoder and classifier are trained, the target encoder is trained to mimic the English encoder with ranking loss and bitext.

In \cref{tab:xnli}, mBERT outperforms one model with bitext training but (as expected) falls short of models with more cross-lingual training information. Interestingly, mBERT and MLM are mostly the same except for the training languages, yet we observe that mBERT under-performs MLM by a large margin. We hypothesize that limiting pretraining to only those languages needed for the downstream task is beneficial. The gap between \citet{artetxe2018massively} and mBERT in XNLI is larger than MLDoc, likely because XNLI is harder.

\insertNerTable

\paragraph{NER} We use \newcite{xie-etal-2018-neural} as a zero-shot cross-lingual transfer baseline, which is state-of-the-art on CoNLL 2002 and 2003. It uses unsupervised bilingual word embeddings \cite{conneau2017word} with a hybrid of a character-level/word-level LSTM, self-attention, and a CRF. Pseudo training data is built by word-to-word translation with an induced dictionary from bilingual word embeddings.

mBERT outperforms a strong baseline by an average of 6.9 points absolute F1 and an 11.8 point absolute improvement in German with a simple one layer 0$^\text{th}$-order CRF as a prediction function (\cref{tab:ner}). A large gap remains when transferring to distantly related languages (e.g. Chinese) compared to a supervised baseline. Further effort should focus on transferring between distantly related languages. In  \cref{sec:exp4} we show that sharing subwords across languages helps transfer.

\insertPosTable

\paragraph{POS} We use \newcite{kim-etal-2017-cross} as a reference. They utilized a small amount of supervision in the target language as well as English supervision so the results are not directly comparable. \cref{tab:pos} shows a large (average) gap between mBERT and Kim et al. Interestingly, mBERT still outperforms \newcite{kim-etal-2017-cross} with 320 sentences in German (de), Polish (pl), Slovak (sk) and Swedish (sv).

\insertParsingTable

\paragraph{Dependency Parsing} We use the best performing model on average in \newcite{ahmad-etal-2019-difficulties} as a zero-shot transfer baseline, i.e. transformer encoder with graph-based parser \cite{dozat2016deep}, and dictionary supervised cross-lingual embeddings \cite{smith2017offline}. Dependency parsers, including \citeauthor{ahmad-etal-2019-difficulties}, assume access to gold POS tags: a cross-lingual representation. We consider two versions of mBERT: with and without gold POS tags. When tags are available, a tag embedding is concatenated with the final output of mBERT.

\cref{tab:parsing} shows that mBERT outperforms the baseline on average by 7.3 point UAS and 0.4 point LAS absolute improvement even without gold POS tags. Note in practice, gold POS tags are not always available, especially for low resource languages.
Interestingly, the LAS of mBERT tends to weaker than the baseline in languages with less word order distance, in other words, more closely related to English.
With the help of gold POS tags, we further observe 1.6 points UAS and 4.7 point LAS absolute improvement on average. It appears that adding gold POS tags, which provide clearer cross-lingual representations, benefit mBERT.

\paragraph{Summary} Across all five tasks, mBERT demonstrate strong (sometimes state-of-the-art) zero-shot cross-lingual performance without any cross-lingual signal. It outperforms cross-lingual embeddings in four tasks. With a small amount of target language supervision and cross-lingual signal, mBERT may improve further; we leave this as future work. In short, mBERT is a surprisingly effective cross-lingual model for many NLP tasks.

\subsection{Question \#2: Does mBERT vary layer-wise?}\label{sec:exp2}

\begin{figure*}[ht]
\centering
\subfloat[][Document classification (ACC)]{
\includegraphics[width=0.82\columnwidth]{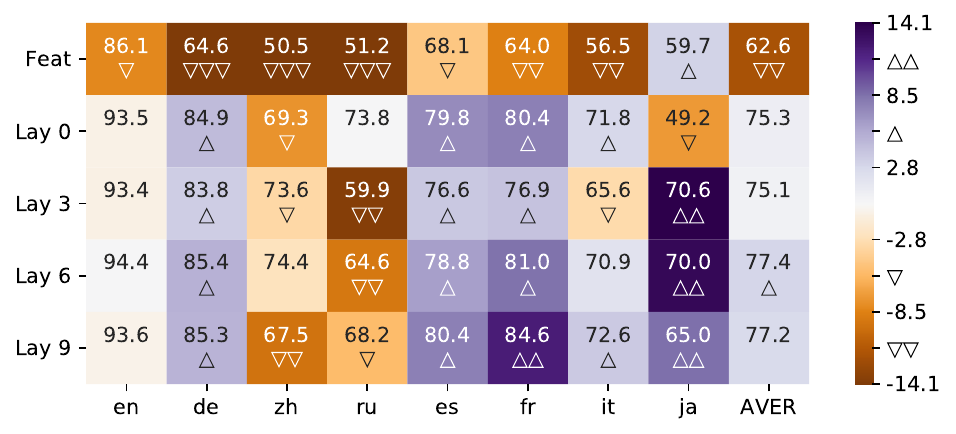}
\label{fig:heatmap-mldoc}}
\subfloat[][Natural language inference (ACC)]{
\includegraphics[width=1.18\columnwidth]{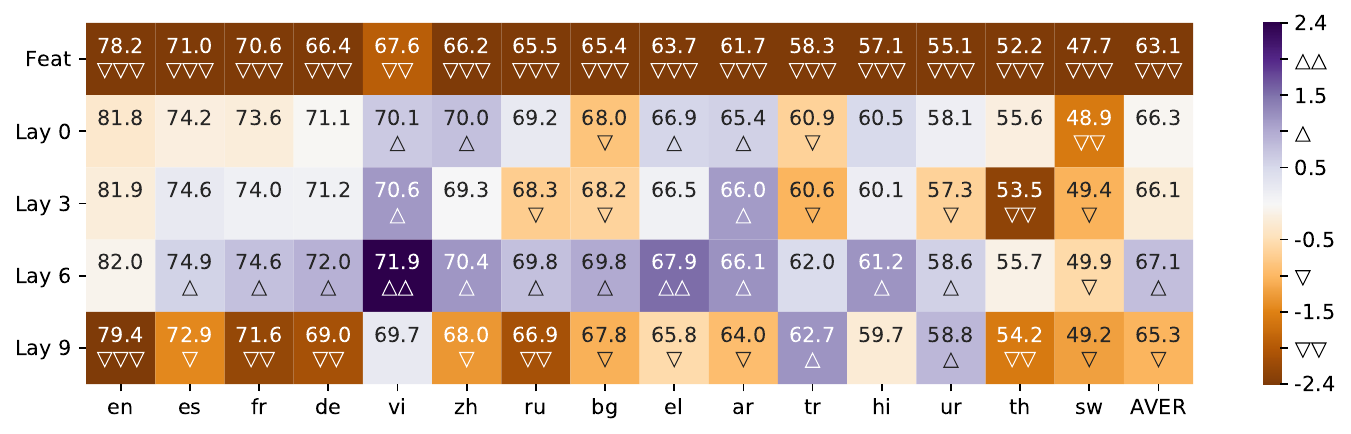}
\label{fig:heatmap-xnli}}
\qquad

\subfloat[][NER (F1)]{
\includegraphics[width=0.67\columnwidth]{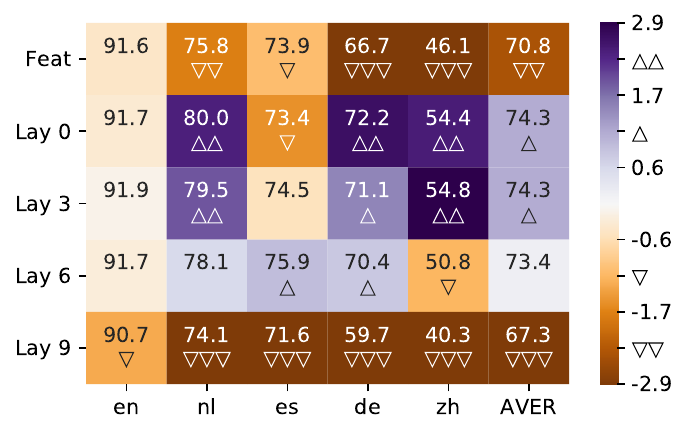}
\label{fig:heatmap-ner}}
\subfloat[][POS tagging (ACC)]{
\includegraphics[width=1.33\columnwidth]{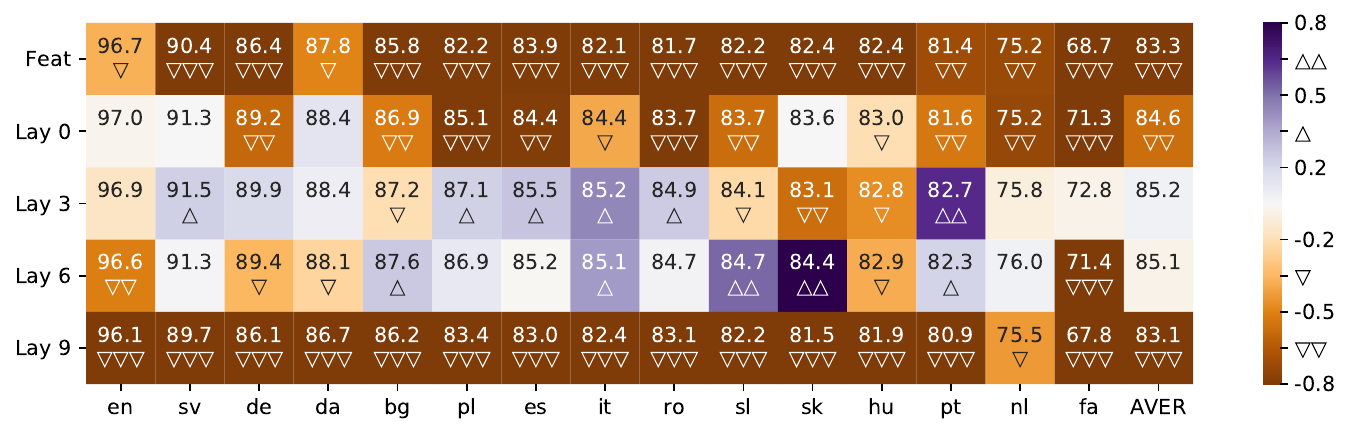}
\label{fig:heatmap-pos}}
\qquad

\subfloat[][Dependency parsing (LAS)]{
\includegraphics[width=2\columnwidth]{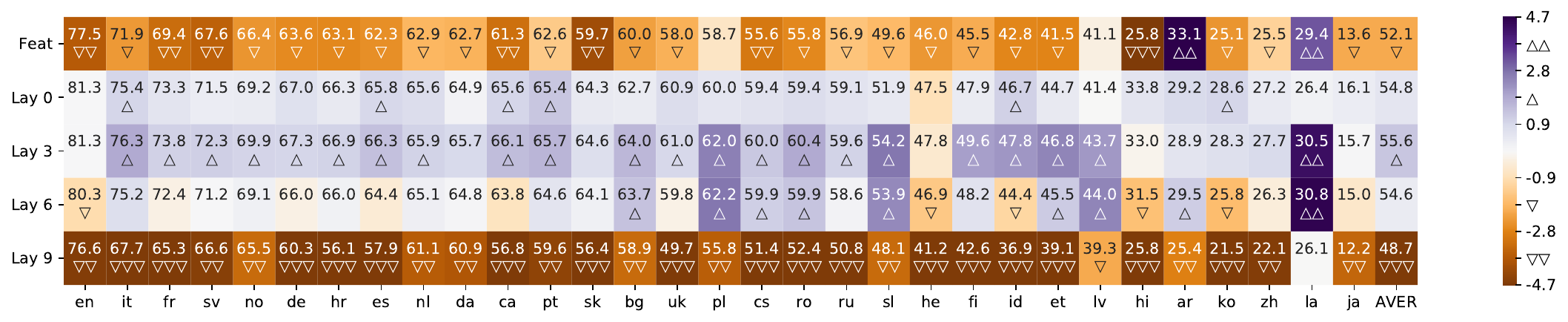}
\label{fig:heatmap-parsing}}

\caption{Performance of different fine-tuning approaches compared with fine-tuning all mBERT parameters. Color denotes absolute difference and number in each entry is the evaluation in the corresponding setting. Languages are sorted by mBERT zero-shot transfer performance. Three downward triangles indicate performance drop more than the legends lower limit.
}\label{fig:heatmap}
\end{figure*}

The goal of a deep neural network is to abstract to higher-order representations as you progress up the hierarchy \cite{yosinski2014transferable}. \newcite{peters-etal-2018-deep} empirically show that for ELMo in English the lower layer is better at syntax while the upper layer is better at semantics. However, it is unclear how different layers affect the quality of cross-lingual representation. For mBERT, we hypothesize a similar generalization across the 13 layers, as well as an abstraction away from a specific language with higher layers. Does the zero-shot transfer performance vary with different layers?

We consider two schemes. First, we follow the feature-based approach of ELMo by taking a learned weighted combination of all 13 layers of mBERT with a two-layer bidirectional LSTM with $d_h$ hidden size (Feat). Note the LSTM is trained from scratch and mBERT is fixed. For sentence and document classification, an additional max-pooling is used to extract a fixed-dimension vector. We train the feature-based approach with Adam and learning rate 1e-3. The batch size is 32. The learning rate is halved whenever the development evaluation does not improve. The training is stopped early when learning rate drop below 1e-5. Second, when fine-tuning mBERT, we fix the bottom $n$ layers ($n$ included) of mBERT, where layer 0 is the input embedding. We consider $n \in \{0, 3, 6, 9\}$.

Freezing the bottom layers of mBERT, in general, improves the performance of mBERT in all five tasks (\cref{fig:heatmap}). For sentence-level tasks like document classification and natural language inference, we observe the largest improvement with $n = 6$. For word-level tasks like NER, POS tagging, and parsing, we observe the largest improvement with $n = 3$. More improvement in under-performing languages is observed.

In each task, the feature-based approach with LSTM under-performs fine-tuning approach. We hypothesize that
initialization from pretraining with lots of languages provides a very good starting point that is hard to beat. Additionally, the LSTM could also be part of the problem. In \citet{ahmad-etal-2019-difficulties} for dependency parsing, an LSTM encoder was worse than a transformer when transferring to languages with high word ordering distance to English.

\subsection{Question \#3: Does mBERT retain language specific information?}\label{sec:exp3}

\begin{figure}[t]
\centering
\includegraphics[width=\columnwidth]{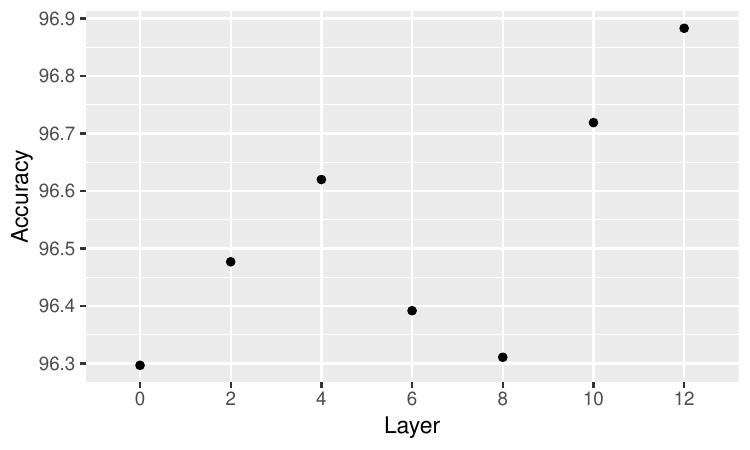}
\caption{Language identification accuracy for different layer of mBERT. layer 0 is the embedding layer and the layer $i > 0$ is output of the i$^\text{th}$ transformer block.}
\label{fig:langid}
\end{figure}

\begin{figure*}[h!]
\centering
\includegraphics[width=2\columnwidth]{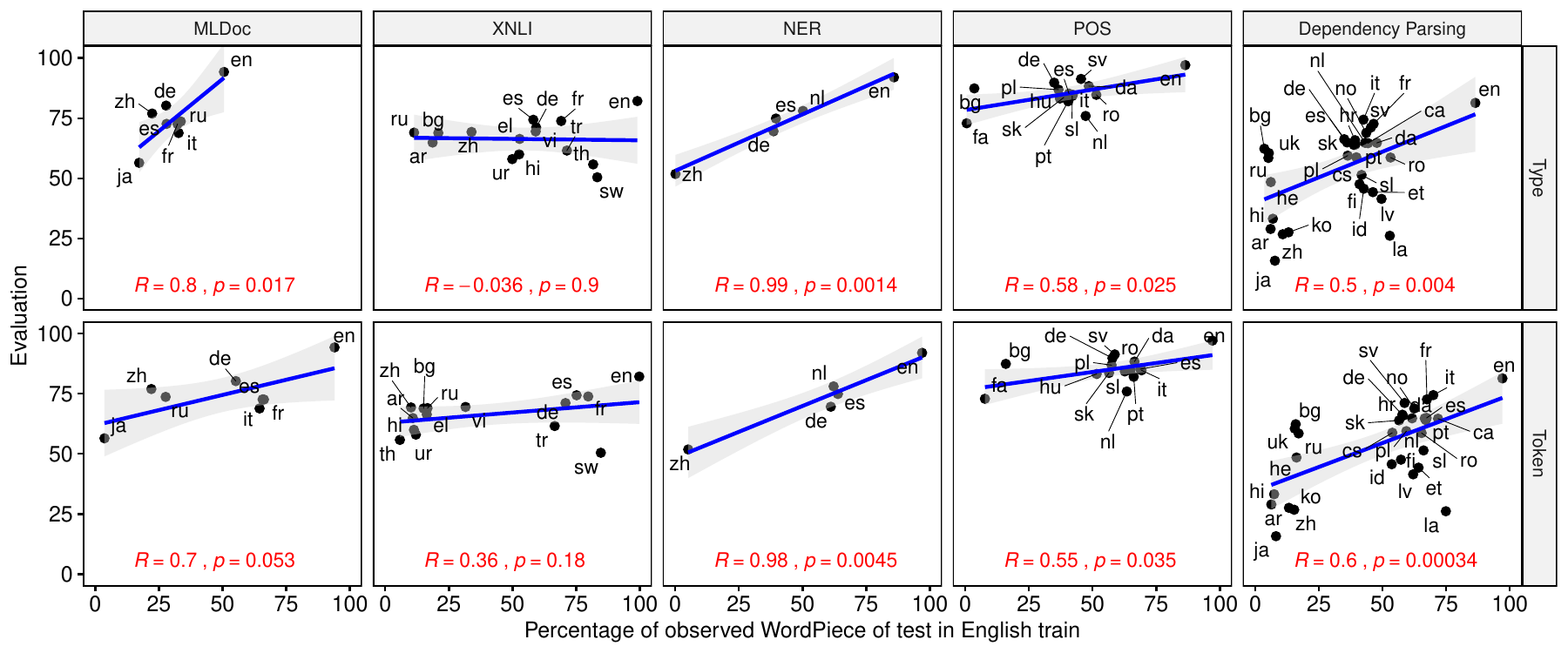}
\caption{Relation between cross-lingual zero-shot transfer performance with mBERT and percentage of observed subwords at both type-level and token-level. Pearson correlation coefficient and $p$-value are shown in red.}\label{fig:corr}
\end{figure*}

mBERT may learn a cross-lingual representation by abstracting away from language-specific information, thus losing the ability to distinguish between languages. We test this by considering language identification: does mBERT retain language-specific information? We use WiLI-2018 \cite{thoma2018wili}, which includes over 200 languages from Wikipedia. We keep only those languages included in mBERT, leaving 99 languages \footnote{Hungarian, Western-Punjabi, Norwegian-Bokmal, and Piedmontese are not covered by WiLI.}.
We take various layers of bag-of-words mBERT representation of the first two sentences of the test paragraph and add a linear classifier with softmax. We fix mBERT and train \textit{only} the classifier the same as the feature-based approach in \cref{sec:exp2}.

All tested layers achieved around 96\% accuracy (\cref{fig:langid}), with no clear difference between layers. This suggests each layer contains language-specific information; surprising given the zero-shot cross-lingual abilities. As mBERT generalizes its representations and creates cross-lingual representations, it maintains language-specific details. This may be encouraged during pretraining since mBERT needs to retain enough language-specific information to perform the cloze task.

\subsection{Question \#4: Does mBERT benefit by sharing subwords across languages?}\label{sec:exp4}

As discussed in \cref{sec:mBERT}, mBERT shares subwords in closely related languages or perhaps in distantly related languages. At training time, the representation of a shared subword is explicitly trained to contain enough information for the cloze task in all languages in which it appears. During fine-tuning for zero-shot cross-lingual transfer, if a subword in the target language test set also appears in the source language training data, the supervision could be leaked to the target language explicitly. However, all subwords interact in a non-interpretable way inside a deep network, 
and subword representations could overfit to the source language and potentially hurt transfer performance. In these experiments, we investigate how sharing subwords across languages effects cross-lingual transfer.

To quantify how many subwords are shared across languages in any task, we assume $V^\text{en}_{\text{train}}$ is the set of all subwords in the English training set, $V^\ell_{\text{test}}$ is the set of all subwords in language $\ell$ test set, and $c^\ell_w$ is the count of subword $w$ in test set of language $\ell$. We then calculate the percentage of observed subwords at type-level $\ptype^\ell$ and token-level $\ptoken^\ell$ for each target language $\ell$.
\begin{align*}
\ptype^\ell &= \frac{|V^\ell_\text{obs}|}{|V^{\ell}_\text{test}|} \times 100 \\
\ptoken^\ell &= \frac{ \sum_{w\in V^\ell_\text{obs}} c^\ell_w } { \sum_{w\in V^\ell_\text{test}} c^\ell_w } \times 100
\end{align*}
where $V^\ell_\text{obs} = V^\text{en}_{\text{train}} \cap V^\ell_{\text{test}}$.

In \cref{fig:corr}, we show the relation between cross-lingual zero-shot transfer performance of mBERT and $\ptype^\ell$ or $\ptoken^\ell$ for all five tasks with Pearson correlation. In four out of five tasks (not XNLI) we observed a strong positive correlation  ($p<0.05$) with a correlation coefficient larger than 0.5. In Indo-European languages, we observed $\ptoken^\ell$ is usually around 50\% to 75\% while $\ptype^\ell$ is usually less than 50\%. This indicates that subwords shared across languages are usually high frequency\footnote{With the data-dependent WordPiece algorithm, subwords that appear in multiple languages with high frequency are more likely to be selected.}. We hypothesize that this could be used as a simple indicator for selecting source language in cross-lingual transfer with mBERT. We leave this for future work.

\section{Discussion}
We show mBERT does well in a cross-lingual zero-shot transfer setting on five different tasks covering a large number of languages. 
It outperforms cross-lingual embeddings, which typically have more cross-lingual supervision. By fixing the bottom layers of mBERT during fine-tuning, we observe further performance gains. Language-specific information is preserved in all layers. Sharing subwords helps cross-lingual transfer; a strong correlation is observed between the percentage of overlapping subwords and transfer performance.

mBERT effectively learns a good multilingual representation with strong cross-lingual zero-shot transfer performance in various tasks. We recommend building future multi-lingual NLP models on top of mBERT or other models pretrained similarly. Even without explicit cross-lingual supervision, these models do very well. As we show with XNLI in \cref{sec:exp1}, while bitext is hard to obtain in low resource settings, a variant of mBERT pretrained with bitext \cite{lample2019cross} shows even stronger performance. Future work could investigate how to use weak supervision to produce a better cross-lingual mBERT, or adapt an already trained model for cross-lingual use. With POS tagging in \cref{sec:exp1}, we show mBERT, in general, under-performs models with a small amount of supervision while \citet{devlin-etal-2019-bert} show that in English NLP tasks, fine-tuning BERT only needs a small amount of data. Future work could investigate when cross-lingual transfer is helpful in NLP tasks of low resource languages. With such strong cross-lingual NLP performance, it would be interesting to prob mBERT from a linguistic perspective in the future.

\bibliography{anthology,main}
\bibliographystyle{acl_natbib}

\cleardoublepage
\appendix

\end{document}